\documentclass[letterpaper, 10 pt, conference]{ieeeconf}  

\IEEEoverridecommandlockouts                              

\usepackage{graphics} 
\usepackage{epsfig} 
\usepackage{amsmath} 
\usepackage{amssymb}  
\usepackage{balance}
\usepackage{url}
\usepackage{color}
\usepackage{booktabs}
\usepackage{subcaption}
\usepackage{multirow}
\usepackage[table]{xcolor}
\usepackage{hyperref}

\usepackage{fancyhdr}
\fancypagestyle{arxivhdr}
{
   \fancyhf{}
   \setlength{\headheight}{15pt} 
\fancyfoot[C]{This paper has been accepted for publication at the\\2026 IEEE/RSJ International Conference on Intelligent Robots and Systems (IROS)}
\fancyhead[C]{\footnotesize Please cite this paper as:\\
Daniel Fusaro, Simone Mosco, Wanmeng Li, and Alberto Pretto, "TravKAN: Fast and Interpretable Nonlinear Traversability Analysis with Kolmogorov-Arnold Networks," 2026 IEEE/RSJ International Conference on Intelligent Robots and Systems (IROS)}
}

\definecolor{cvprblue}{rgb}{0.21,0.49,0.74}

\newcommand{\new}[1]{\textcolor{black}{#1}}

\title{\LARGE \bf
TravKAN: Fast and Interpretable Nonlinear Traversability Analysis with Kolmogorov-Arnold Networks
}

\author{Daniel Fusaro, Simone Mosco, Wanmeng Li, and Alberto Pretto 
\thanks{\{fusarodani, moscosimon, liwanmeng, alberto.pretto\}@dei.unipd.it
}
\thanks{Department of Information Engineering, University of Padova, Italy. %
}
}

\begin{document}

\maketitle
\thispagestyle{empty}
\pagestyle{empty}
\thispagestyle{arxivhdr}

\begin{abstract}
Traversability analysis is a fundamental capability for autonomous mobile robots operating in unstructured environments.
While modern machine learning approaches such as deep neural networks and gradient-boosted trees achieve strong predictive performance, they lack interpretability and provide limited insight into the underlying terrain-robot interaction dynamics.
In this paper, we propose TravKAN, a Kolmogorov-Arnold Network-based framework for fast, scalable, and interpretable traversability estimation.
TravKAN represents multivariate decision functions through compositions of learnable univariate functions, enabling compact architectures and symbolic extraction of analytic expressions after training. In addition, we introduce a novel set of handcrafted features derived from the reflectivity channel of LiDAR sensors.
To the best of our knowledge, reflectivity has not been systematically exploited for handcrafted traversability descriptors, despite its potential to capture material and surface properties complementary to geometric cues.
We evaluate TravKAN on public, real-world urban and off-road datasets and compare it against strong baselines. TravKAN achieves strong performance across all metrics, outperforming conventional deep models and approaching the performance of XGBoost.
TravKAN-Lite, i.e., TravKAN's symbolic representation, reveals meaningful nonlinear feature interactions and provides a compact, deployment-friendly, and fast analytic model.
Ablation studies further show the robustness of our method to architectural variations and quantify the contribution of the proposed reflectivity-based features.
These properties make TravKAN attractive for robotic systems requiring transparency, real-time computational efficiency, and interpretability in safety-critical decision-making.
\end{abstract}

\section{INTRODUCTION}

\begin{figure}
    \centering
    \includegraphics[width=\columnwidth]{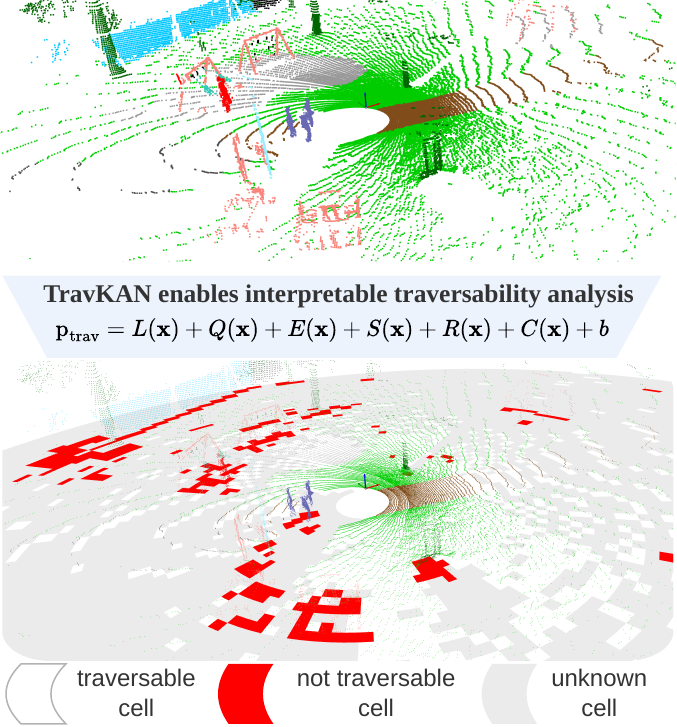}
    \caption{By learning nonlinear transformations directly on edges and parameterizing them with B-splines, captures complex feature interactions while maintaining a compact architecture.
    After training, the learned model can be symbolically extracted into an analytic expression, referred to here as TravKAN-Lite, that reveals the underlying decision function in terms of explicit feature contributions and nonlinearities.
    This formulation is computationally efficient for deployment (extremely reduced runtime and $O(1)$ computational complexity) and allows for deeper insights into the model's behavior.}
    \label{fig:teaser}
\end{figure}

Autonomous mobile robots operating in natural and semi-structured environments must continuously assess terrain traversability to ensure safe and efficient navigation. Traversability analysis is the task of estimating the traversability score of the environment surrounding a vehicle in order to ensure safe navigation~\cite{bellone2018tits}.
Accurate traversability estimation directly impacts path planning, energy consumption, mobility risk, and mission success.
As robotic platforms are increasingly deployed in off-road and urban-edge environments, reliable and scalable traversability analysis becomes essential.

Recent advances in machine learning~(ML) have significantly improved data-driven traversability analysis.
Deep neural networks such as multilayer perceptrons~(MLP) and Transformer-based~\cite{vaswani2017neurips} models can capture complex nonlinear relationships in high-dimensional features.
Similarly, gradient-boosted decision trees, particularly XGBoost~\cite{chen2016kddm}, have demonstrated strong performance on tabular data tasks.
However, despite their predictive power, these approaches suffer from a critical drawback: limited interpretability.
Neural networks act as black boxes, offering little insight into feature interactions, while tree ensembles produce large, difficult-to-analyze rule sets.
In safety-critical robotic systems, understanding why a terrain patch is classified as traversable or not is often as important as the prediction itself.

In this work, we investigate whether Kolmogorov-Arnold Networks~(KANs)~\cite{liu2024kan} can provide an explainable alternative for traversability analysis.
KANs parameterize multivariate nonlinear functions as compositions of learnable univariate functions, enabling compact representations and symbolic extraction after training.
This structure makes them particularly appealing for robotics applications where transparency, compactness, and analytical structure are desirable.
Unlike conventional deep neural networks, KANs yield explicit symbolic expressions approximating the learned decision function, providing insight into nonlinear feature interactions.
We propose TravKAN, an interpretable KAN-based framework for scalable terrain traversability analysis.
Based on handcrafted features derived from LiDAR point clouds, including a novel set of reflectivity-based descriptors, TravKAN models nonlinear dependencies while maintaining a compact architecture suitable for deployment.
After training, the learned model can be converted into an analytic expression, offering both interpretability and fast, deterministic inference cost, as shown in~Fig.~\ref{fig:teaser}. 

We evaluate our approach on three public, large-scale, real-world datasets: SemanticKITTI (urban driving)~\cite{behley2019iccv}, nuScenes (multi-city, multi-weather)~\cite{caesar2020cvpr}, and EastPark (off-road and urban scenarios)~\cite{eastpark}.
Experimental results show that TravKAN achieves competitive performance across all metrics, outperforming conventional deep models and approaching the performance of XGBoost.
Importantly, this competitive performance is achieved while retaining the ability to extract compact symbolic representations of the learned decision boundary.

The contributions of this paper are fivefold:
\begin{itemize}
    \item We introduce TravKAN, a Kolmogorov-Arnold Network-based framework for interpretable nonlinear traversability modeling.
    \item We introduce a new set of handcrafted features based on the reflectivity channel of the LiDAR sensor.
    \item We provide a comprehensive experimental comparison against deep learning, SVM, and gradient-boosted baselines on urban and off-road, public, real-world datasets.
    \item We demonstrate that TravKAN enables symbolic extraction of the learned model, offering transparency and deployment advantages while maintaining competitive predictive performance.
    \item We carry out an ablation study demonstrating the robustness to design change of TravKAN, and the information contribution of the new handcrafted features.
\end{itemize}

By bridging nonlinear function approximation and symbolic interpretability, TravKAN offers a promising direction toward transparent and scalable decision-making in robotic navigation systems. The code can be found at~\href{https://github.com/Bender97/travKAN}{https://github.com/Bender97/travKAN}.

\section{Related Work}
Traversability analysis has been extensively studied in the robotics community, with a wide range of approaches proposed over the years.
The concept of traversability depends on the robot's size and capabilities, as well as the terrain's geometric and material properties.
For an unmanned ground vehicle, traversability is often defined as the ability to safely and effectively navigate over a terrain patch without getting stuck, tipping over, or damaging the robot~\cite{fusaro2023ras}.
Hence, traversability can be modeled as a binary classification problem, where the goal is to predict whether a given terrain patch is traversable or not based on sensor data.
Typical sensor modalities used for traversability analysis include LiDAR point clouds, RGB images, and RGB-D data.

LiDAR point clouds are particularly informative for traversability analysis, as they provide rich geometric information about the terrain surface.
LiDAR-based traversability analysis methods typically extract features from the point cloud, such as height, roughness, slope, and curvature, and then apply machine learning models to predict traversability at the point or patch level, with a patch often defined as a local neighborhood of points or a discretized cell in a grid representation.
Methods can be categorized into three main groups: machine learning-based, deep learning-based, and reinforcement learning-based methods.

\subsection{Machine Learning-based Methods}

Kernel-based support vector machine (SVM) methods are widely used.
The common approach is to manually design feature descriptors that capture relevant geometric and/or radiometric properties of the terrain, and then train an SVM classifier to predict traversability based on these features.
\cite{bellone2018tits} proposes a normalized descriptor enriched with geometric and color information, subsequently detecting road traversability in point clouds acquired in outdoor environments.
Similarly, \cite{narvaez2018terrain} analyzes different terrains using a texture descriptor that combines color and depth information.
\cite{zhou2012self} utilizes Fuzzy~SVM and features extracted from terrain in forest environments. 
\cite{fusaro2023ras} and~\cite{li2024cisram} propose a pyramidal 3D feature fusion approach that aggregates multi-scale geometric and radiometric features and applies multiple SVM classifiers for traversability estimation.
Differently, \cite{4161557,7139749} are based on Bayesian inference. \cite{4399610} exploits the K-nearest neighbors algorithm to obtain the traversability map.

\subsection{Deep Learning-based Methods}

Driven by recent deep learning advancements, an increasing number of 3D point cloud processing models have been proposed.
A solution to the traversability analysis problem is to exploit a proxy task, like semantic segmentation, and then map the predicted semantic classes to traversability labels.
For example, methods like LSK3DNet~\cite{feng2024cvpr}, Point Transformer v3~\cite{wu2024cvpr}, WaffleAndRange~\cite{fusaro2024exploiting}, and 2DPASS~\cite{yan20222dpass} perform semantic segmentation on point clouds obtaining state-of-the-art performance on large-scale datasets.
Nonetheless, semantic segmentation is not the optimal solution for traversability analysis, as it requires a large amount of annotated data and may not capture the specific geometric and material properties relevant for traversability.
This is why many works have focused on directly modeling traversability without relying on semantic segmentation as an intermediate step.

Considering the success of 2D convolutional neural networks~(CNNs), one of the most intuitive approaches involves projection, where 3D point clouds are projected into a range view \cite{8967762},\cite{9561171},\cite{8793495} or a bird-eye view \cite{9156460}.
Then, 2D CNNs are utilized to extract feature representations in these image planes, followed by view feature fusion to generate the final output representations.

Another alternative approach is based on 3D voxelization \cite{8953494,8579059, ruetz2024foresttrav}, followed by 3D convolutions.
These methods have demonstrated good performance, but are subject to nontrivial loss of geometric detail due to quantization on the voxel grid.
To enable deep network structures to directly process point clouds, pioneering works such as PointNet \cite{8099499} and PointNet++ \cite{NIPS2017_d8bf84be} consider point clouds as sets embedded in continuous space.
Point-based models \cite{9156466},\cite{9010002} can benefit from this idea, and several approaches have been developed.

Conversely, generative adversarial networks (GANs) have been explored for traversability analysis.
GANs have the advantage of being able to learn complex data distributions and generate realistic samples, which can be useful for data augmentation and improving model robustness.
A variety of semi-supervised GANs~\cite{8594031,8624332,8750823} and unsupervised GANs~\cite{7989540} have been proposed.

\subsection{Reinforcement Learning-based Methods}
Modern reinforcement learning-based methods rely on trial-and-error interactions to make predictions.
Most models \cite{9149720},\cite{9196879},\cite{s21030796},\cite{8468643} tend to experiment in elaborate simulated environments, generating enormous overhead in the process of producing terrain environments that are close to real samples.
Some of the latest work \cite{9345970} uses an autonomous data labeling strategy.
This enables autonomous mobile systems to obtain a self-supervised traversability map of the real environment.
However, collecting and labeling data from dangerous behaviors, such as collisions, remain challenging. 
A few recent works \cite{9354889}, \cite{9304721} have explored applying deep inverse reinforcement learning to traversability analysis.
In the context of self-supervised learning, auto-exploration methods like~\cite{tang2019iros} have been proposed to enable robots to autonomously explore and learn about their environment.
They need a large amount of expert demonstration data to estimate the underlying reward function.
However, they have only been evaluated on simple scenes and have not been tested in complex environments.\\

In our work, we focus on supervised learning-based methods, using handcrafted features extracted from LiDAR point clouds as input to a KAN-based model.
Despite the strong performance of deep learning and reinforcement learning-based methods, we argue that the interpretability, fast runtime, and deployment advantages of KANs make them a compelling alternative for traversability analysis in real-world robotic systems.

\section{Our Approach}
We aim to develop a fast, interpretable, and scalable traversability analysis KAN-based model using handcrafted features extracted from LiDAR point clouds.
We formulate traversability analysis as a cell-wise binary classification problem, where each cell corresponds to a discretized patch of the terrain.
We base our feature extraction on the handcrafted features proposed in~\cite{li2024cisram}, which capture geometric and remission properties of the terrain, and we extend this feature set to include novel reflectivity-based descriptors.
The compact architecture allows for easy deployment on robotic platforms while maintaining competitive performance.

\subsection{Problem Formulation}

Let $\mathbf{P} = \{\mathbf{p_i}\}_{i=1}^{N}$ denote a LiDAR point cloud of $N$ points acquired in the sensor coordinate frame. 
Each point $\mathbf{p_i}$ is represented as a K-dimensional measurement, including (depending on the sensor used to acquire the point cloud) $x_i, y_i, z_i \in \mathbb{R}$ as the 3D Cartesian coordinates of the point in the sensor frame, $I_i \in \mathbb{R}$ as the corresponding return remission, and $R_i \in \mathbb{R}$ as the corresponding reflectivity.

Rather than predicting traversability at the individual point level, we model it as a spatial property of the ground surface.
To obtain a structured representation aligned with LiDAR sampling geometry, the point cloud is projected onto the ground plane and discretized in polar coordinates $(r, \theta)$, with $ r \in [0, +\infty), \theta \in [0, 2 \pi)$. This induces a polar grid:
\begin{equation}
\mathcal{G} = \{C_k\}_{k=1}^{M},
\end{equation}
where each cell $C_k$ aggregates a subset of~$\mathbf{P}$:
\begin{equation}
    C_k = \{\mathbf{p_i} \in \mathbf{P} \mid  (r_i, \theta_i) \in \text{bin}_k \}.
\end{equation}

The polar grid representation provides a resolution that grows naturally with distance and is structurally consistent with LiDAR sampling geometry.

For each polar cell $C_k$, we compute an aggregated handcrafted feature vector $\mathbf{v_k} \in \mathbb{R}^d$, following the handcrafted features described in~\cite{li2024cisram}.



Each polar cell is then independently represented by a set of geometric and radiometric descriptors, described in~\ref{sec:reflectivity}.
The final feature vector for each cell is composed of:
\begin{equation}
    \mathbf{v_k} = [\mathbf{v_k}^{\text{geom}},\mathbf{v_k}^{\text{remission}},\mathbf{v_k}^{\text{reflectivity}}],
\end{equation}
where geometric, remission, and reflectivity descriptors are concatenated into a unified representation.

Traversability estimation is therefore formulated as a cell-wise binary classification problem. We learn a mapping: 
$f_\theta : \mathbb{R}^d \rightarrow [0, 1]$,
such that $\hat{y}_k = f_\theta(\mathbf{v_k})$ represents the probability that cell $C_k$ is traversable.
Ground-truth labels are defined per cell: $y_k \in \{0, 1\}$.

\subsection{KAN Formulation}
To approximate $f_\theta$, we adopt Kolmogorov--Arnold Networks (KANs)~\cite{liu2024kan}, which parameterize nonlinear transformations directly on edges rather than nodes and enable symbolic
extraction of analytic expressions after training.

Let $f : \mathbb{R}^{n} \rightarrow \mathbb{R}^{m}$ denote a mapping from two feature spaces. 
Standard neural networks such as~MLPs parameterize $f$ as compositions of affine maps and fixed nonlinearities:
\begin{equation}
f(x) = W_L \sigma\!\left(W_{L-1} \sigma(\cdots \sigma(W_1 x))\right),
\end{equation}
where the activation function $\sigma$ is fixed (e.g., ReLU, GELU). 
Expressivity is therefore concentrated in weight matrices, while nonlinearities remain globally shared and static.

KANs invert this allocation of capacity by learning nonlinear functions directly on edges. 
Motivated by the Kolmogorov--Arnold superposition theorem, which guarantees that continuous multivariate functions admit decompositions into sums of univariate functions composed with linear projections, KANs parameterize mappings as superpositions of learned one-dimensional transformations.

\subsection{KAN Layer Formulation}
A KAN layer mapping $\mathbb{R}^{n} \to \mathbb{R}^{m}$ is defined as
\begin{equation} \label{eq:kan_layer}
z_j = \sum_{i=1}^{n} \phi_{ij}(a_{ij} x_i + b_{ij}), \quad j = 1, \dots, m,
\end{equation}
where:
\begin{itemize}
    \item $a_{ij}, b_{ij} \in \mathbb{R}$ are learnable affine scalars,
    \item $\phi_{ij} : \mathbb{R} \rightarrow \mathbb{R}$ are learnable univariate functions.
\end{itemize}
Nonlinearity is thus \emph{edge-based} rather than node-based.

Stacking layers yields $f(x) = \mathcal{L}_L \circ \dots \circ \mathcal{L}_1(x),$ where each layer follows Eq.~\ref{eq:kan_layer}.

\begin{figure*}[ht]
    \centering
    \includegraphics[width=\linewidth]{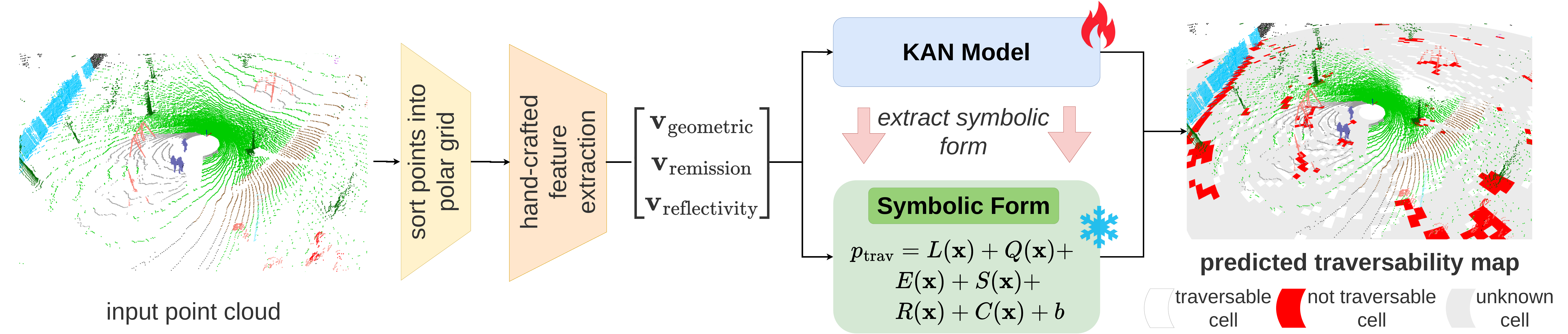}
    \caption{By learning nonlinear transformations directly on edges and parameterizing them with B-splines, TravKAN captures complex feature interactions while maintaining a compact architecture.
    After training, the learned model can be symbolically extracted into an analytic expression that reveals the underlying decision function in terms of explicit feature contributions and nonlinearities.
    This formulation is computationally efficient for deployment (extremely reduced runtime and $O(1)$ computational complexity) and allows for deeper insights into the model's behavior.}
    \label{fig:archi}
\end{figure*}

\subsection{Spline Parameterization}

Each univariate function (that KANs aim to learn) $\phi_{ij}$ is parameterized using a B-spline basis expansion over $[-1,1]$:
\begin{equation}
\phi_{ij}(t) = \sum_{k=1}^{G} c_{ijk} B_k(t),
\end{equation}
where:
\begin{itemize}
    \item $\{B_k\}_{k=1}^{G}$ are order-$r$ B-spline basis functions,
    \item $c_{ijk} \in \mathbb{R}$ are learnable coefficients,
    \item $G$ is the grid resolution.
\end{itemize}

KAN models utilize B-splines because they hold compact support and $C^{r-1}$ smoothness, allowing highly localized parameter adaptability without triggering global representation collapse.
Hence, each edge implements a smooth, locally controlled nonlinear operator.

Since B-splines form a universal basis on compact intervals, each $\phi_{ij}$ can approximate any sufficiently smooth univariate function. 
Layer stacking enables interaction terms via composition, yielding universal approximation.
It's this parameterization that allows for direct symbolic extraction of the learned functions after training, as each $\phi_{ij}$ can be explicitly represented as a weighted sum of known basis functions, such as sinusoids, exponentials, or polynomials.

Capacity is controlled by the hyperparameter $G$, the grid resolution.
After an initial training with a fixed $G$, for sufficiently smooth functions, a grid refinement yields systematic approximation error decay (for details, refer to~\cite{liu2024kan}).
Hence, unlike width scaling in MLPs or Transformers, increasing~$G$ during training refines local functional resolution without altering network topology, often leading to more stable optimization and better generalization.

Each learned edge function is explicitly accessible, 
allowing direct visualization of learned nonlinear transformations.

\subsection{KAN Model}

We propose TravKAN, a KAN-based model for traversability estimation. Fig.~\ref{fig:archi} shows the architecture of TravKAN.
For each polar cell $C_k$, we compute a handcrafted feature vector $\mathbf{v_k} \in \mathbb{R}^d$ that captures geometric, remission, and reflectivity properties of the terrain patch, as done in~\cite{li2024cisram}. 
TravKAN is implemented as a shallow KAN with a single hidden layer. The network maps the input feature space to the output traversability classes through the following layer dimensions:
$[n_\text{in}, n_\text{hidden}, n_\text{out}] = [d, 24, 2],$
 where $d$ denotes the dimensionality of the handcrafted feature vector.

Despite being shallow, TravKAN can capture complex nonlinear interactions through the edge-based spline parameterization.

\subsection{Reflectivity Features} \label{sec:reflectivity}
To complement the geometric and remission-based features, we introduce a novel set of reflectivity-based descriptors computed from the reflectivity channel of the LiDAR sensor.
While remission and reflectivity are related radiometric measurements, they are physically distinct.
Remission values are influenced by sensor-specific characteristics and acquisition geometry, whereas reflectivity (often referred to as calibrated intensity) provides a more intrinsic measure of surface reflectance, typically calibrated to reduce range and angle dependency.
To the best of our knowledge, reflectivity has not been systematically exploited for handcrafted traversability modeling, and remains heavily underexplored in perception.
In~\cite{viswanath2025reflectivity}, the authors claim that adding reflectivity as input enhances the performance of deep learning LiDAR segmentation models, but they do not explore the potential of reflectivity for handcrafted feature design.

We compute reflectivity-based descriptors analogously to the remission-based features introduced in~\cite{li2024cisram}, including minimum, maximum, mean, variance, dispersion of reflectivity within the cell and radial distributional measures.
These reflectivity features complement geometric cues by providing material-related information, which is particularly informative in urban and off-road scenarios where terrain classes may exhibit similar geometry but different surface properties (e.g., asphalt vs. gravel, grass vs. soil).

We evaluate the contribution of these features through an ablation study in Section~\ref{sec:ablation}.

\section{Experiments}

We evaluate TravKAN, our proposed fast and interpretable KAN-based traversability analysis model, on three datasets and compare it against strong baselines, including MLPs, Transformer encoders, SVM-based methods, and XGBoost.

\subsection{Datasets}

We benchmark on three large-scale LiDAR datasets:

\begin{itemize}
    \item \textbf{SemanticKITTI} (urban driving, dense annotations),
    \item \textbf{nuScenes} (multi-city, multi-weather),
    \item \textbf{EastPark} (off-road, natural park and urban scenarios).
\end{itemize}

Traversability labels are derived from semantic annotations by mapping drivable and flat terrain classes to the positive class and obstacles, vegetation, vehicles, and vertical structures to the negative class. 
This produces large-scale binary supervision across heterogeneous environments. We followed the ground truth extraction protocol proposed in~\cite{li2024cisram}.

All datasets are split following official training/validation/test protocols.
For EastPark, we train on sequences 00 and 03, validate on sequences 01 and 04, and evaluate on sequences 02 and 05.
The evaluation on SemanticKITTI and NuScenes is done on the standard validation set.
We train and validate on the original training set by splitting it into training and validation sets using an 80\%-20\% random split.

\subsection{Baselines}

We compare TravKAN against:

\begin{itemize}
    \item \textbf{MLP}: Multi-Layer Perceptron consisting of stacked linear layers. We used layer sizes $[d, 32, 64, 128, 256, 64, 2]$,
    \item \textbf{MLP++}: MLP in which each linear layer (except the last one) is followed by LeakyReLU activation and batch norm. We used same layer sizes,
    \item \textbf{TransformerEncoder}: Transformer encoder with 6 layers, 4 attention heads, and hidden dimension of 1024 and no positional encoding, since the input features are unordered; an initial Linear layer maps the input features to the hidden dimension, and a final Linear layer maps the output of the last encoder layer to the 2 output classes,
    \item \textbf{P-SVM}: Pyramidal 3D feature fusion with SVM classifier, as proposed in~\cite{fusaro2023ras} and extended in~\cite{li2024cisram},
    \item \textbf{XGBoost}: Gradient-boosted decision tree with 3000 trees, maximum depth of 6, and learning rate of 0.1.
\end{itemize}

\subsection{Implementation Details}
All methods are trained using the same handcrafted feature set, extracted following~\cite{li2024cisram} and extended with reflectivity-based descriptors as described in Section~\ref{sec:reflectivity} (only for the EastPark dataset).
All deep learning-based models are implemented in PyTorch and trained on a single NVIDIA Titan RTX.
The CPU used for all experiments is an Intel(R) Core(TM) i7-7700K CPU @ 4.20GHz.

All methods are trained for a maximum of 100 epochs with early stopping based on no improvement in validation loss for 10 consecutive epochs.
We trained all methods using Cross-Entropy loss.
We used the AdamW optimizer, with an initial learning rate of 0.005 and weight decay of 1e-5, and a ReduceLROnPlateau scheduler that reduces the learning rate by a factor of 0.5 if validation loss does not improve for 5 epochs.
Batch size is fixed across methods to 30000 feature vectors. 
P-SVM is trained using the implementation provided by the authors in~\cite{li2024cisram}, with hyperparameters set according to their recommendations and using the extended handcrafted features as the other methods for a fair comparison.

The number of input features depends on the dataset, as the feature set is extended with reflectivity-based descriptors only for EastPark.
Thus, for EastPark $d = 36$, while for SemanticKITTI and NuScenes $d = 29$.

We trained TravKAN in a 3-step procedure, following the training protocol proposed in~\cite{liu2024kan}:
\begin{itemize}
    \item Step 1: Initial training with grid resolution $G=8$ for 50 epochs (early stopping applies if validation loss does not improve for 10 epochs),
    \item Step 2: Grid refinement to $G=40$ and training for 30 epochs with learning rate linearly decreased with coefficient 0.125 (with the same early stopping),
    \item Step 3: Final grid refinement to $G=48$ and training for 20 epochs with learning rate linearly decreased with coefficient 0.125 and early stopping).
\end{itemize}
All hyperparameters are selected based on validation performance, and the final evaluation is performed on the test set using the best model checkpoint obtained during training.

\subsection{Evaluation Metrics}

We report Balanced Accuracy (Bal. Acc.), mIoU, F1-score, Area Under the ROC Curve (ROC AUC), and Area Under the Precision-Recall Curve (PR AUC).
Bal. Acc. is reported to mitigate class imbalance, while ROC AUC and PR AUC evaluate ranking robustness across thresholds.
mIoU and F1 score are reported for completeness, but are threshold-dependent and thus secondary to the ranking metrics.

\subsection{Main Results}

\begin{table}[t]
\centering
\caption{Performance comparison on the EastPark dataset.}
\label{tab:eastpark_results}
\resizebox{1.0\columnwidth}{!}{%
\begin{tabular}{lccccc}
\toprule
\multirow{2}{*}{\textbf{Method}} 
& \multirow{2}{*}{\shortstack{\textbf{Bal. Acc.} \\ $\uparrow$}} 
& \multirow{2}{*}{\shortstack{\textbf{mIoU} \\ $\uparrow$}} 
& \multirow{2}{*}{\shortstack{\textbf{F1} \\ $\uparrow$}} 
& \textbf{ROC} 
& \textbf{PR} \\
& & & & AUC $\uparrow$ & AUC $\uparrow$ \\
\midrule
MLP & 87.6 & \underline{80.3} & \underline{88.8} & 89.8 & 94.1 \\
MLP++ & \textbf{87.9} & 77.8 & 87.2 & \underline{92.2} & \underline{95.6} \\
TransformerEncoder & \underline{87.9} & 79.4 & 88.2 & 91.9 & 95.4 \\    
P-SVM & 87.6 & \textbf{82.0} & \textbf{90.0} & 90.7 & 94.7 \\
XGBoost & 87.4 & 79.5 & 88.3 & 88.6 & 92.2 \\
\rowcolor{cvprblue!15} \textbf{Ours (TravKan)} & 87.8 & 77.6 & 87.0 & \textbf{93.1} & \textbf{96.5} \\
\rowcolor{cvprblue!15} \textbf{TravKan-Lite} & 85.6 & 77.5 & 86.9 & 89.0 & 93.9 \\
\bottomrule
\end{tabular}
}
\end{table}

\begin{table}[t]
\centering
\caption{Perf. comparison on the SemanticKITTI dataset.}
\label{tab:kitti_results}
\resizebox{1.0\columnwidth}{!}{%
\begin{tabular}{lccccc}
\toprule
\multirow{2}{*}{\textbf{Method}} 
& \multirow{2}{*}{\shortstack{\textbf{Bal. Acc.} \\ $\uparrow$}} 
& \multirow{2}{*}{\shortstack{\textbf{mIoU} \\ $\uparrow$}} 
& \multirow{2}{*}{\shortstack{\textbf{F1} \\ $\uparrow$}} 
& \textbf{ROC} 
& \textbf{PR} \\
& & & & AUC $\uparrow$ & AUC $\uparrow$ \\
\midrule
MLP & 84.0 & 67.6 & 80.6 & 92.1 & 83.2 \\
MLP++ & \textbf{93.9} & \textbf{87.9} & \textbf{93.5} & 98.1 & 96.6 \\
TransformerEncoder & 89.7 & 78.8 & 88.1 & 95.8 & 92.5 \\
P-SVM & 91.2 & 87.1 & 93.1 & 96.3 & \underline{97.0} \\
XGBoost & 93.4 & 87.3 & \underline{93.2 }& \underline{98.3} & 96.9 \\
\rowcolor{cvprblue!15} \textbf{Ours (TravKan)} & \underline{93.5} & \underline{86.2} & 92.5 & \textbf{98.4} & \textbf{97.1} \\
\rowcolor{cvprblue!15} \textbf{TravKan-Lite} & 86.3 & 71.2 & 83.1 & 89.0 & 69.8 \\
\bottomrule
\end{tabular}
}
\vspace{-1.5em}

\end{table}

\begin{table}[t]
\centering
\caption{Perf. comparison on the NuScenes dataset.}
\label{tab:nuscenes_results}
\resizebox{1.0\columnwidth}{!}{%
\begin{tabular}{lccccc}
\toprule
\multirow{2}{*}{\textbf{Method}} 
& \multirow{2}{*}{\shortstack{\textbf{Bal. Acc.} \\ $\uparrow$}} 
& \multirow{2}{*}{\shortstack{\textbf{mIoU} \\ $\uparrow$}} 
& \multirow{2}{*}{\shortstack{\textbf{F1} \\ $\uparrow$}} 
& \textbf{ROC} 
& \textbf{PR} \\
& & & & AUC $\uparrow$ & AUC $\uparrow$ \\
\midrule
MLP & 93.1 & 87.0 & 93.1 & 97.4 & 95.6 \\
MLP++ & \textbf{95.8} & \textbf{91.9} & \textbf{95.8} & 99.0 & 98.7 \\
TransformerEncoder & 83.1 & 66.4 & 79.7 & 90.8 & 82.1 \\
P-SVM & 93.4 & 89.9 & 93.2 & 97.9 & 96.9 \\
XGBoost & 95.6 & 91.6 & 95.6 & \textbf{99.1} & \textbf{98.9} \\
\rowcolor{cvprblue!15} \textbf{Ours (TravKan)} & \underline{95.7} & \underline{91.7} & \underline{95.7} & \textbf{99.1} & \textbf{98.9} \\
\rowcolor{cvprblue!15} \textbf{TravKan-Lite} & 89.4 & 80.2 & 89.1 & 95.0 & 91.3 \\
\bottomrule
\end{tabular}
}
\end{table}

Table~\ref{tab:eastpark_results} reports results on EastPark.
\new{TravKan achieves the strongest ranking performance, obtaining the best ROC AUC (93.1\%) and PR AUC (96.5\%), indicating superior discrimination capability and robustness on this imbalanced classification task. In particular, the improvement in PR AUC over the strongest baseline (MLP++, 95.6\%) demonstrates a more reliable identification of the positive class across different decision thresholds.
TravKan's balanced accuracy (87.8\%) is comparable to the other methods, differing by less than 0.2 percentage points from the top result. Likewise, its mIoU (77.6\%) and F1 score (87.0\%) are slightly lower than those of the P-SVM baseline, which achieves the best values for these metrics.
Overall, the results indicate that TravKan offers the best trade-off between threshold-independent discriminative performance and competitive threshold-dependent accuracy, making it particularly suitable for deployment in real-world robotic applications where the operating threshold may vary depending on the desired precision--recall trade-off.}
We refer to the symbolic approximated formulation of TravKAN as Lite-TravKAN, which still achieves competitive performance despite having substantially low runtime and memory requirements.
On SemanticKITTI (Table~\ref{tab:kitti_results}), TravKAN demonstrates strong large-scale generalization.
Compared to XGBoost, TravKAN obtains better Bal. Acc., ROC AUC and PR AUC with a slightly lower mIoU.
Compared to MLP++, it obtains a comparable Bal. Acc. but higher ranking metrics (ROC AUC and PR AUC).

TravKAN scales effectively to larger datasets and higher separability regimes, matching or exceeding tree-based methods.
On NuScenes (Table~\ref{tab:nuscenes_results}), TravKAN achieves competitive performance, trailing MLP++ by only 0.1 Bal. Acc., and is on par with XGBoost in terms of ranking metrics (ROC AUC and PR AUC).
Given that NuScenes exhibits different sensor characteristics and environmental conditions, this result highlights cross-domain robustness.
On the SemanticKITTI and NuScenes datasets, Lite-TravKAN also achieves competitive performance, even outperforming MLP or TransformerEncoder, respectively.

\section {Discussion and Interpretability}
TravKAN is consistently competitive with XGBoost across all benchmarks.
Compared to deep baselines such as MLP and Transformer encoders, TravKAN exhibits stable performance across datasets with different sensing conditions and environmental characteristics.
High ROC AUC and PR AUC values indicate that TravKAN produces well-calibrated confidence scores, which is essential for robotics applications involving risk-aware navigation, adaptive thresholding, and deployment in safety-critical systems.

While XGBoost achieves slightly higher performance on certain datasets, its decision process consists of hundreds of piecewise-constant tree splits, making global interpretation difficult.
A key advantage of TravKAN lies in its ability to extract a closed-form symbolic approximation of the learned decision function.
For the EastPark dataset, symbolic extraction yields an analytic expression that can be written in this compact form:
\begin{equation}
\hat{y}
=
L(\mathbf{x})+
Q(\mathbf{x})+
E(\mathbf{x})+
S(\mathbf{x})+
R(\mathbf{x})+
C(\mathbf{x})+
b,
\label{eq:travkan_symbolic}
\end{equation}

where $\mathbf{x} \in \mathbb{R}^d$ denotes the handcrafted feature vector, $L(\mathbf{x})$, $Q(\mathbf{x})$, $E(\mathbf{x})$,
$S(\mathbf{x})$, $R(\mathbf{x})$, and $C(\mathbf{x})$
denote the learned linear, quadratic, exponential,
trigonometric (sine and cosine), rational, and composite nonlinear interaction terms, respectively, and $b$ is the learned bias.
We report the full expanded symbolic expression for the EastPark model in the Github repository. 

\new{Examination of the learned expression provides insight into the model's decision process by explicitly exposing both the contribution of individual features and their nonlinear interactions. The largest coefficients are consistently associated with height-related statistics, particularly the average, variance, standard deviation, and extrema of the local elevation ($z_{\text{ave}}, z_{\text{var}}, z_{\text{std}}, z_{\text{min}}, z_{\text{max}}$), together with the average intensity, indicating that these descriptors dominate the traversability prediction.
In contrast, geometric features such as the surface normal ($n_{vz}$), unevenness, angle, and surface density mainly appear within nonlinear basis functions, suggesting that they act as local correction terms under specific terrain conditions rather than contributing uniformly across the feature space.
Overall, the symbolic model indicates that traversability is primarily driven by local terrain morphology, while geometric and radiometric descriptors refine the prediction through nonlinear interactions.}

We report in Table~\ref{tab:complexity} parameter counts and runtimes, considering both CPU and GPU implementations.
Runtime is computed considering the average number of polar cells per frame in the test set ($\sim$1100 polar cells) excluding the shared feature extraction time ($\sim$6\,ms), which is the same for all methods.
In Table~\ref{tab:complexity}, we refer to Lite-TravKAN as the symbolic extracted version of TravKAN, where the learned model is approximated by a symbolic expression composed of linear and nonlinear analytic functions including trigonometric, exponential, and polynomial terms as in Eq.~\ref{eq:travkan_symbolic}.
TravKAN achieves a significant reduction in parameter count compared to deep baselines (MLP and Transformer), while maintaining competitive performance.
The model size of TravKAN is 2 orders of magnitude smaller than Transformer, and 3 orders of magnitude smaller than P-SVM and XGBoost.
Lite-TravKAN achieves an additional reduction of 2 orders of magnitude compared to TravKAN.
The CPU runtime of TravKAN is comparable to P-SVM and slightly higher than XGBoost, but significantly higher than MLP and Transformer.
A similar trend is observed for GPU inference.
But Lite-TravKAN is substantially faster than all methods, with an average runtime of 0.16\,ms per frame on CPU, making it suitable for real-time deployment on resource-constrained robotic platforms.

\begin{table}[t]
\centering
\caption{Model complexity and deployment comparison.
Runtime is reported per frame, considering the avg. number of polar cells in the test set of Eastpark (1100 polar cells).
}
\label{tab:complexity}
\resizebox{1.0\columnwidth}{!}{%
\begin{tabular}{lcccc}
\toprule

\multirow{2}{*}{\shortstack{\textbf{Method}}} 
& \multirow{2}{*}{\shortstack{\textbf{\#Params} }} 
& \multirow{2}{*}{\shortstack{\textbf{Model} \\ \textbf{Size}}} 
& \multirow{2}{*}{\shortstack{\textbf{CPU} \\ \textbf{Time (ms)}}} 
& \multirow{2}{*}{\shortstack{\textbf{GPU} \\ \textbf{Time (ms)}}} \\
\\

\midrule
MLP & 61K & 244\,KB & 0.9\,ms & 0.4\,ms \\
MLP++ & 62K & 260\,KB & 1.4\,ms & 0.7\,ms \\
TransformerEncoder & 2M & 7.6\,MB & 5.3\,ms & 4.3\,ms \\
P-SVM & 30K & 25\,MB & 50\,ms & / \\
XGBoost & 560K & 68\,MB & 25\,ms & / \\
\rowcolor{cvprblue!15} \textbf{TravKAN (Ours)} & \underline{10K} & 72\,KB & 71\,ms & 60\,ms \\
\rowcolor{cvprblue!15} \textbf{Lite-TravKAN (Ours)} & $\mathbf{<}$ \textbf{1K} & $\mathbf{<}$ \textbf{1KB} & 0.16\,ms & / \\
\bottomrule
\end{tabular}
}
\end{table}

\section{Ablation Study}
\label{sec:ablation}

\textit{Architecture Variation.}
\new{Table~\ref{tab:ablation-architecture} reports the results of the ablation study evaluating the influence of the KAN hidden dimension and the initial grid resolution. Overall, the proposed architecture exhibits stable performance across all tested configurations, with PR AUC from 94.8 to 96.5, and balanced accuracy from 85.4 to 88.2. No monotonic relationship is observed between either hyperparameter and the evaluation metrics, indicating that the two design choices interact with each other. In particular, the optimal grid resolution depends on the selected hidden dimension, and vice versa. The highest PR AUC (96.5) are achieved by the 24–Grid8 configuration, whereas the highest balanced accuracy of 88.2 is obtained with 16–Grid8. However, several configurations achieve nearly identical performance, with differences below 0.5\% in balanced accuracy, suggesting that the proposed method is relatively insensitive to these hyperparameters. This robustness indicates that no single configuration consistently dominates across all evaluation metrics, allowing the final architecture to be selected based on a trade-off between predictive performance and computational complexity.}

\textit{Reflectivity Features.}
To assess the contribution of the newly introduced reflectivity-based features, we perform an ablation study on the EastPark dataset by comparing the full TravKAN model with a variant that excludes reflectivity features (geom+remission) and another variant that excludes both remission and reflectivity features (geom).
The results are reported in Table~\ref{tab:ablation-reflectivity}.
As shown, the inclusion of reflectivity features leads to a consistent improvement in all AUC-based metrics and balanced accuracy, which indicates that reflectivity provides complementary information that enhances the model's ability to distinguish between traversable and non-traversable terrain.

\begin{table}[t]
\centering
\caption{Ablation study on stability of our TravKAN under architecture variation on the EastPark dataset (hidden dimension by row, grid size by column).}
\label{tab:ablation-architecture}

\begin{subtable}[t]{0.48\linewidth}
\centering
\caption{\textbf{Balanced Accuracy (\%)}}

\resizebox{1\columnwidth}{!}{%
\begin{tabular}{c|cccc}
\toprule
\textbf{Ours} & \textbf{8} & \textbf{16} & \textbf{24} & \textbf{30} \\
\midrule
8  & 87.3 & 87.8 & 86.2 & 88.0 \\
16 & \textbf{88.2} & 85.8 & 86.4 & 85.4 \\
24 & 87.8 & 87.6 & 86.5 & 87.0 \\
30 & 87.1 & 87.9 & \underline{88.0} & 87.5 \\
\bottomrule
\end{tabular}
}
\end{subtable}
\begin{subtable}[t]{0.48\columnwidth}
\centering
\caption{\textbf{PR AUC (\%)}}
\resizebox{0.98\columnwidth}{!}{%
\begin{tabular}{c|cccc}
\toprule
\textbf{Ours} & \textbf{8} & \textbf{16} & \textbf{24} & \textbf{30} \\
\midrule
8  & 96.2 & 96.1 & 96.1 & 96.3 \\
16 & 96.4 & 95.1 & 96.2 & 94.8 \\
24 & \textbf{96.5} & 96.0 & 96.0 & 96.1 \\
30 & 96.3 & \underline{96.5} & 96.4 & 96.3 \\
\bottomrule
\end{tabular}}
\end{subtable}

\end{table}

\begin{table}[t]
\centering
\caption{Ablation study on the EastPark dataset to evaluate the contribution of reflectivity-based features.}
\label{tab:ablation-reflectivity}
\resizebox{1.0\columnwidth}{!}{%

\begin{tabular}{lccccc}
\toprule
\multirow{2}{*}{\textbf{Method}} 
& \multirow{2}{*}{\shortstack{\textbf{Bal. Acc.} \\ $\uparrow$}} 
& \multirow{2}{*}{\shortstack{\textbf{mIoU} \\ $\uparrow$}} 
& \multirow{2}{*}{\shortstack{\textbf{F1} \\ $\uparrow$}} 
& \textbf{ROC} 
& \textbf{PR} \\
& & & & AUC $\uparrow$ & AUC $\uparrow$ \\
\midrule
geom only & 87.4 & \underline{78.3} & \underline{87.5} & \underline{92.7} & \underline{96.4} \\
geom+remission & \underline{87.5} & \textbf{78.4} & \textbf{87.5} & 92.6 & 96.3 \\
\rowcolor{cvprblue!15} \textbf{Full TravKan} & \textbf{87.8} & 77.6 & 87.0 & \textbf{93.1} & \textbf{96.5} \\
\bottomrule
\end{tabular}
}
\end{table}

\section{Conclusion}
We introduced TravKAN, a novel KAN-based model for fast LiDAR-based traversability estimation that achieves competitive performance while providing symbolic interpretability and computational efficiency.
It consistently achieves competitive performance while providing the strongest ranking capability as measured by ROC AUC and PR AUC across three large-scale datasets, demonstrating robustness and generalization.
The ability to extract a closed-form symbolic expression of the learned decision function offers unique insights into feature importance and interactions, which is crucial for safety-critical applications in robotics.
An interesting future work will be to explore the potential of KAN models to discover new symbolic descriptors directly from raw correlation matrices, rather than relying on handcrafted features.


\bibliographystyle{IEEEtran}
\bibliography{glorified,bib} 

\end{document}